\pdfoutput=1

\documentclass[11pt]{article}

\usepackage[]{acl}

\usepackage{times}
\usepackage{latexsym}

\usepackage[T1]{fontenc}

\usepackage[utf8]{inputenc}

\usepackage{microtype}

\usepackage{inconsolata}
\usepackage{graphicx}
\usepackage{xcolor}
\usepackage{amsmath}
\usepackage{makecell}
\usepackage{subcaption}

\title{Sentence-level Media Bias Analysis with Event Relation Graph}

\author{Yuanyuan Lei and Ruihong Huang\\
        Department of Computer Science and Engineering\\
        Texas A\&M University, College Station, TX\\
        \texttt{\{yuanyuan, huangrh\}@tamu.edu}}

\begin{document}
\maketitle
\begin{abstract}

Media outlets are becoming more partisan and polarized nowadays. In this paper, we identify media bias at the sentence level, and pinpoint bias sentences that intend to sway readers' opinions. As bias sentences are often expressed in a neutral and factual way, considering broader context outside a sentence can help reveal the bias. In particular, we observe that events in a bias sentence need to be understood in associations with other events in the document. Therefore, we propose to construct an event relation graph to explicitly reason about event-event relations for sentence-level bias identification. The designed event relation graph consists of events as nodes and four common types of event relations: coreference, temporal, causal, and subevent relations. Then, we incorporate event relation graph for bias sentences identification in two steps: an event-aware language model is built to inject the events and event relations knowledge into the basic language model via soft labels; further, a relation-aware graph attention network is designed to update sentence embedding with events and event relations information based on hard labels. Experiments on two benchmark datasets demonstrate that our approach with the aid of event relation graph improves both precision and recall of bias sentence identification \footnote{The code and data link: https://github.com/yuanyuanlei-nlp/sentence\_level\_media\_bias\_naacl\_2024}.

\end{abstract}

\section{Introduction}

\begin{figure*}[t]
  \centering
  \includegraphics[width = 6.3in]{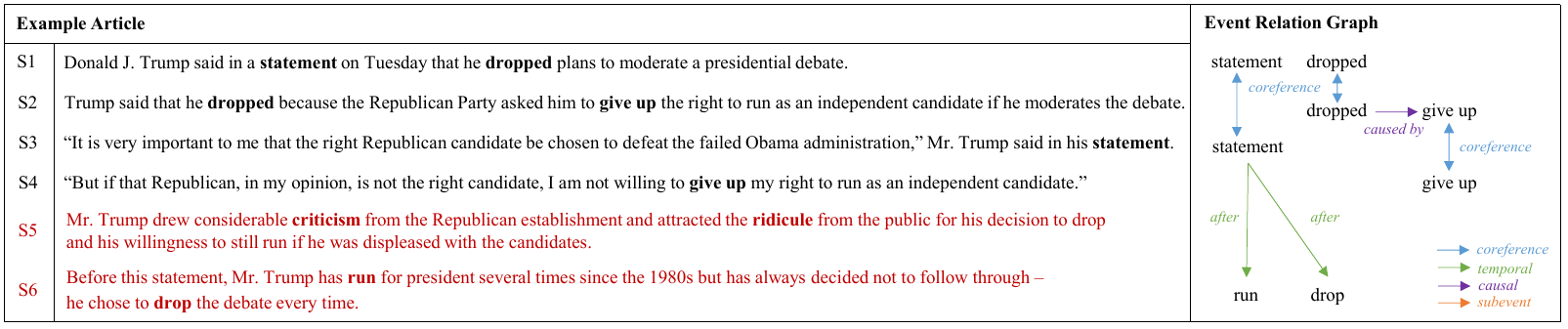}
  \caption{An example article containing bias sentences, and its corresponding event relation graph. Bias sentences are highlighted in \textcolor{red}{red}. Events words are shown in \textbf{bold} text. Event relation graph consists of events as nodes and four types of event relations: coreference, temporal, causal, and subevent relation.}
  \label{introduction_example}
\end{figure*}

Media bias refers to the ideological leaning of the journalists within news reporting \cite{lichter2017theories, gentzkow2006media}. News media plays an important role not only in providing information, but also in shaping public opinions \cite{van1991publics, van1995power, stromback2012media, 10008998}. Media outlets are exhibiting growing partisanship and polarization, with a greater potential to influence public opinions and interfere democratic process \cite{prior2013media, wilson2020polarization, lei-etal-2022-sentence}. Thus, developing sophisticated models to detect media bias is important and necessary.

While media bias detection has received growing research interests, the majority of prior work detect media bias at the article level \cite{baly-etal-2020-detect, kiesel2019semeval}. Nevertheless, each sentence within an article serves for different purposes in narrating a news story. It is important to pinpoint those bias sentences that aim to implant ideological bias and manipulate public opinions. Bias sentence, as interpreted by \citet{10.1086/499414, 10.1111/j.1460-2466.2006.00336.x, RePEc:nbr:nberwo:9295}, is the content providing supportive or background information to shift pubic opinions in an ideological direction, though that may be done via information selective inclusion or omission as well as overt ideological language. This paper aims to identify media bias at sentence level, and develop computational model to detect bias sentences.

Identifying sentence-level media bias still remains a challenging task \cite{vargas-etal-2023-predicting}, especially considering its subtle and implicit nature. Take the example article in Figure \ref{introduction_example} as an instance, while the first bias sentence S5 contains overt ideological language and may be easily identified, the second bias sentence S6 looks completely neutral and factual, and a model merely examining the sentences in isolation without considering broader context may not readily reveal such bias cases. But it becomes clear that S6 carries bias if we contrast S6 (\textit{Trump always decided not to follow through presidential selections}) with what Trump said in the statement (\textit{his willingness to still run if he was displeased with the candidates}). Interestingly, it appears that the author hinted on the relatedness between the historical events described in S6 and the \textit{statement} event by explicitly stating a temporal relation between them. To further elaborate, Trump claimed a causal relation between events \textit{drop plans to moderate a debate} and \textit{give up right as an independent candidate}, revealing the historical events in S6 propels the reader to rethink whether the stated cause is indeed the real reason, given his consistent past behavior of \textit{dropping presidential election several times since 1980s}.

While it remains difficult to fully encode the intricate reasoning process used for recognizing sentence-level bias, inspired by the observed importance of understanding events in the context of other related events, we propose to construct an event relation graph and connect events in a document with four common types of event relations: coreference, temporal, causal and subevent relations. The event relation graph explicitly captures event-level content structures and enhances the comprehension of events within their interrelated context. We employ this event relation graph to guide the bias sentence detector and engage its attention on significant event-event relations.

Moreover, we propose to integrate the event relation graph into bias sentences identification in two ways. Firstly, an event-aware language model is trained using the soft labels derived from the event relation graph, thereby injecting the knowledge of events and event-event relations into the language model. Secondly, a relation-aware graph attention network is designed to encode the event relation graph based on hard labels, and update sentence embedding with events and event relations information. The utilization of both soft labels and hard labels of the event relation graph are complementary: soft labels provide nuanced probabilistic information and necessitate the basic language model to recognize that nodes represent events and links represent event relations, while hard labels facilitate updating sentence embeddings with interconnected events and event relation embeddings. Experiments on two benchmark datasets demonstrate the effectiveness of our approach based on the event relation graph, yielding performance gains on both precision and recall. The ablation study confirms the necessity and synergy of leveraging both soft labels and hard labels derived from the event relation graph. Our main contributions are summarized as:

\begin{itemize}
    \item We firstly observe that interpreting events in association with other events in a document is critical for identifying bias sentences.
    \item We propose a new framework to incorporate the event relation graph as extra guidance for sentence-level media bias identification.
    \item We effectively improve the F1 score for bias sentences identification by 5.78\% on BASIL dataset and 12.86\% on BiasedSents dataset.
\end{itemize}

\section{Related Work}

\noindent\textbf{Source-level Media Bias} attracted research attention in prior years. The work in this area assumed all the articles and journalists within one media outlet share the same political ideology. Prior work such as \citet{groseclose2005measure, gentzkow2010drives} measures source-level media bias via citation patterns or consumer political preferences. \citet{budak2016fair} implemented a crowd-sourcing approach to rate the political slant of media sources. \citet{baly-etal-2018-predicting} combined text-based methods with social media behaviors to create a bias classification system.

\vspace{6pt}

\noindent\textbf{Article-level Media Bias} has been researched for years \cite{wang-2017-liar, lei-huang-2023-identifying}. Early work used text-based approach \cite{sapiro2019examining}. \citet{chen-etal-2020-detecting} developed Gaussian mixture model to incorporate fine-grained bias information. \citet{baly-etal-2020-detect} collected a large-scale dataset and provided manual labels, showing that only three percentage of articles have a different ideology label from their source's. \citet{liu-etal-2022-politics} pre-trained a political domain language model. Different from previous work on source-level or article-level media bias, we aim to identify media bias at the fine-grained sentence-level, and pinpoint the content that can illuminate and explain the overall article bias.

\noindent\textbf{Sentence-level Media Bias} has a relatively short research history \cite{lei-huang-2022-shot, lei-huang-2023-discourse, vargas-etal-2023-predicting}. The initial work in this field is \citet{fan-etal-2019-plain}, where they annotate sentence-level media bias within political news document. Another work \cite{lim-etal-2020-annotating} also annotate media bias at sentence level while considering the article context. A discourse structure method was developed by \citet{lei-etal-2022-sentence}, in which the discourse structures knowledge were distilled into bias sentence identification via a knowledge distillation framework.  Different from \citet{lei-etal-2022-sentence}, our event relation graph interprets news narrative from the perspective of event reporting, and enables constructing event-level discourse relations for every possible pair of sentences, fostering a comprehensive understanding of the broader context.

\noindent\textbf{Event Graph} was introduced by \citet{li-etal-2020-connecting} where they presented a graph-based structure connecting events and entities nodes through event argument role relations. This event graph was deployed in several down-stream applications, such as sentence fusion \cite{yuan-etal-2021-event}, story generation \cite{chen-etal-2021-graphplan}, and misinformation detection \cite{wu-etal-2022-cross}. The event graph constructed in their work comprises entity-entity links and event-entity links via event argument roles, while lacks event-event relations. Differently, our proposed event relation graph focuses on events and establishes event interconnections through four types of event-event relations.

\noindent\textbf{Event and Event Relations} were researched for decades. Previous work explored event identification and event extraction \cite{du-cardie-2020-event, liu-etal-2020-event}. There are four common relations between events: coreference \cite{zeng-etal-2020-event, barhom-etal-2019-revisiting}, temporal \cite{ning-etal-2018-multi, han-etal-2019-joint}, causal \cite{zuo-etal-2021-improving, gao-etal-2019-modeling}, and subevent relations \cite{aldawsari-finlayson-2019-detecting, lai-etal-2022-multilingual-subevent}. While each type of relation was previously studied in isolation, a recent work \cite{wang-etal-2022-maven} provided the first large-scale corpus with all four relations annotated. Instead of modeling each event relation separately, we aim to develop a model that unifies all four relations for comprehensive discourse analysis.

\section{Event Relation Graph}

The event relation graph can incorporate broader contextual information outside the single sentence, and thus help reveal the underlying bias inside the sentence. Hence, we propose to create an event relation graph for each article. In this section, we explain the components, training process, and construction process of the event relation graph.

\subsection{Event Relation Graph Components}

The event relation graph consists of events as nodes and four types of event relations as links. An event refers to an occurrence or action reported in the article. Coreference relation informs us whether the two events designate the same event. Temporal relation represents the chronological order. Instead of only reflecting the existence of temporal order, we classify temporal relation into three detailed categories for deeper narrative understanding: \textit{before}, \textit{after}, and \textit{overlap}. Causal relation shows the causality or precondition relation between events, and we classify causal relation into two specific types: \textit{causes} or \textit{caused by}. Subevent relation recognizes containment or subordination relation between events, and is categorized into \textit{contains} and \textit{contained by}. Hence, the designed event relation graph not only identifies the presence of each relation, but also offers more detailed and in-depth interconnection information.

\begin{figure*}[t]
  \centering
  \includegraphics[width = 6.3in]{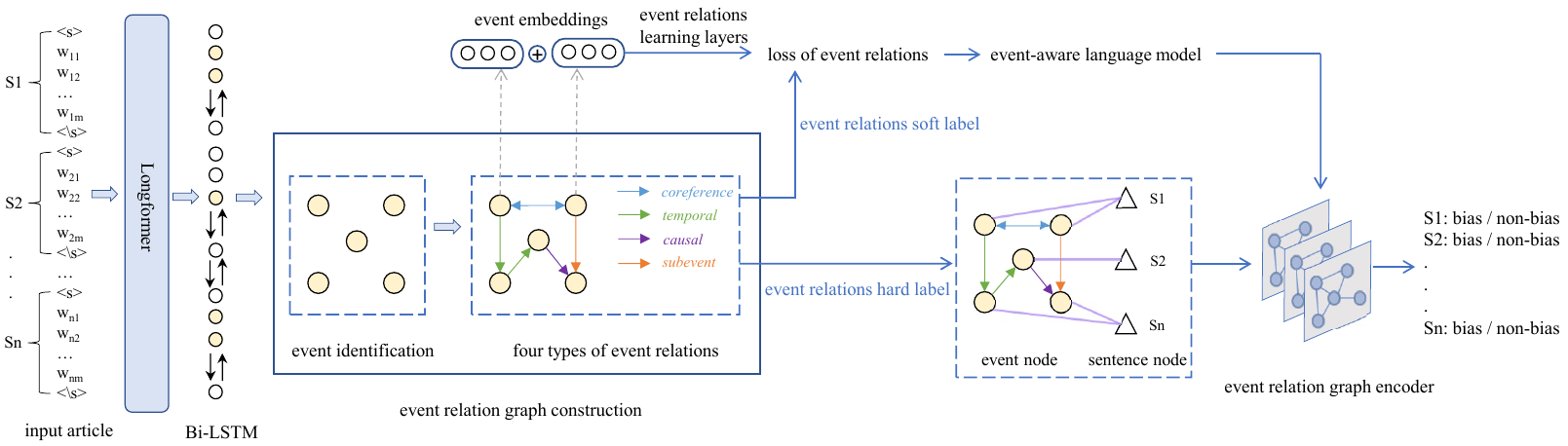}
  \caption{An illustration of sentence-level media bias identification based on event relation graph}
  \label{methodology_figure}
\end{figure*}

\subsection{Event Relation Graph Training}

The event relation graph is trained on the general-domain MAVEN-ERE dataset \cite{wang-etal-2022-maven}. Following the previous work \cite{yao-etal-2020-weakly}, we form the training event pairs in the natural textual order, where the former event in each pair is the precedent event mentioned in the text. In terms of temporal relations, the dataset also annotates the time expressions such as date or time. Considering our event relation graph focuses on events, we solely retain annotations between events. We further process the \textit{before} annotation in this way: keep the \textit{before} label if the annotated event pair aligns with the natural textual order, or reverse the event pair and assign the \textit{after} label if not. The \textit{simultaneous}, \textit{overlap}, \textit{begins-on}, \textit{ends-on}, \textit{contains} annotations are combined into the \textit{overlap} category in our event relation graph. In terms of causal relations, we maintain the \textit{causes} label if the natural textual order is followed, or assign the \textit{caused by} label if not. Similar process for the \textit{contains} and \textit{contained by} labels within subevent relations. Since events and four event relations interact with each other to form a cohesive narrative framework, we adopt joint learning framework \cite{wang-etal-2022-maven} to train these components collaboratively. The training process results in an event identifier and four event relations extractors.

\subsection{Event Relation Graph Construction}

Given a candidate news article, an event relation graph is created. During the construction process, we first identify which word triggers an event by using the trained event identifier to generate the predicted probability for each word:
\begin{equation}
    \small
    P_i^{event} = (p_i^{non-event}, p_i^{event})
\end{equation}
where \small$i=1,...,N$\normalsize, \small$N$ \normalsize is the number of words in the article, and $p_i^{event}$ is the probability of $i$-th word designating an event.

Next, we establish all possible event pairs based on the identified events and extract the four relations between them. To accomplish this, we employ the trained four relations extractors to generate the predicted probabilities for each event pair $(event_i, event_j)$ as follows:
\begin{equation}
    \small
    P_{i, j}^{corefer} = (p_{i, j}^{non-corefer}, p_{i, j}^{corefer})
\end{equation}
\begin{equation}
    \small
    P_{i, j}^{temp} = (p_{i, j}^{non-temp}, p_{i, j}^{before}, p_{i, j}^{after}, p_{i, j}^{overlap})
\end{equation}
\begin{equation}
    \small
    P_{i, j}^{causal} = (p_{i, j}^{non-causal}, p_{i, j}^{cause}, p_{i, j}^{caused-by})
\end{equation}
\begin{equation}
    \small
    P_{i, j}^{subevent} = (p_{i, j}^{non-subevent}, p_{i, j}^{contain}, p_{i, j}^{contained-by})
\end{equation}
where $i$ and $j$ denote the index of two events. \small$p_{i, j}^{corefer}$ \normalsize is the probability of two events corefer with each other. \small$p_{i, j}^{non-temp}$ \normalsize indicates the probability of no temporal relation existing between them, and \small$p_{i, j}^{before}, p_{i, j}^{after}, p_{i, j}^{overlap}$ \normalsize represents the probability of the relations \textit{before}, \textit{after}, \textit{overlap} correspondingly. Similar denotations for the causal and subevent relations.

In the latter bias sentences identification process, the predicted probabilities in equation (1)-(5) are incorporated as the \textit{soft labels} for the constructed event relation graph. The \textit{hard labels} for the events and four relations are derived by applying the $argmax$ function on the soft labels.

\section{Bias Sentences Identification}

The event relation graph is incorporated into bias sentences identification within two consecutive steps, as illustrated in Figure \ref{methodology_figure}. Firstly, an event-aware language model is developed to inject the events and event relations knowledge into the basic language model using the \textit{soft labels}. Secondly, a relation-aware graph attention network is devised to encode the event relation graph based on \textit{hard labels}, and update sentence embeddings with events and event relations information.

\subsection{Event-Aware Language Model}

The event-aware language model aims to inject the events and event relations knowledge from the soft labels into the basic language model. Since the news articles are usually long, we utilize the Longformer \cite{beltagy2020longformer} as the basic language model to encode the entire article and derive each word embedding. We also add an extra layer of Bi-LSTM \cite{huang2015bidirectional} on top to capture the contextual information.

To enhance the basic language model with the event relation graph knowledge, we construct an event learning layer on top of word embeddings $(w_1, w_2,...,w_N)$ to learn the events knowledge, and also build four relations learning layers on top of event pair embeddings $(e_i \oplus e_j)$ to learn the four event relations respectively:
\begin{equation}
    \small
    \begin{split}
        Q_i^{event} & = (q_i^{non-event}, q_i^{event}) \\
        & = softmax(W_2(W_1 w_i + b_1) + b_2)
    \end{split}
\end{equation}
\begin{equation}
    \small
    \begin{split}
        Q_{i, j}^{corefer} & = (q_{i, j}^{non-corefer}, q_{i, j}^{corefer}) \\
        & = softmax(W_4(W_3 (e_i \oplus e_j) + b_3) + b_4)
    \end{split}
\end{equation}
\begin{equation}
    \small
    \begin{split}
        Q_{i, j}^{temp} & = (q_{i, j}^{non-temp}, q_{i, j}^{before}, q_{i, j}^{after}, q_{i, j}^{overlap}) \\
        & = softmax(W_6(W_5 (e_i \oplus e_j) + b_5) + b_6)
    \end{split}
\end{equation}
\begin{equation}
    \small
    \begin{split}
        Q_{i, j}^{causal} & = (q_{i, j}^{non-causal}, q_{i, j}^{cause}, q_{i, j}^{caused-by}) \\
        & = softmax(W_8(W_7 (e_i \oplus e_j) + b_7) + b_8)
    \end{split}
\end{equation}
\begin{equation}
    \small
    \begin{split}
        Q_{i, j}^{subevent} & = (q_{i, j}^{non-subevent}, q_{i, j}^{contain}, q_{i, j}^{contained-by})\\
        & = softmax(W_{10}(W_9 (e_i \oplus e_j) + b_9) + b_{10})
    \end{split}
\end{equation}
where \small$W_1,...,W_{10}, b_1,...,b_{10}$ \normalsize are trainable parameters in the learning layers. \small$(e_i \oplus e_j)$ \normalsize is the embedding for the event pair $(event_i, event_j)$ by concatenating the two events embeddings together. \small $Q_i^{event}$ \normalsize, and \small $Q_{i, j}^{corefer}, Q_{i, j}^{temp}, Q_{i, j}^{causal}, Q_{i, j}^{subevent}$ \normalsize are the learned probabilities for events and four event relations of the basic language model.

The soft labels generated by the event relation graph \footnotesize$P_i^{event}, P_{i, j}^{corefer}, P_{i, j}^{temp}, P_{i, j}^{causal}, P_{i, j}^{subevent}$ \normalsize contain informative knowledge of events and event relations, thereby are referenced as the target learning materials. By minimizing the cross entropy loss between the learned probability and the target probability, the events and event relations knowledge within the event relation graph can be injected into the basic language model:
\begin{equation}
    \small
    Loss_{event} = -\sum_{i=1}^NP_i^{event}\log(Q_i^{event})
\end{equation}
\begin{equation}
    \small
    Loss_{r} = -\sum_{i,j}P_{i, j}^{r}\log(Q_{i, j}^{r})
\end{equation}
where \small$r \in \{corefer, temp, causal, subevent\}$ \normalsize represents the four event relations. The overall learning objective for training the event-aware language model based on soft labels is to minimize the cumulative loss for learning each component:
\begin{equation}
    \small
    \begin{split}
        Loss_{soft} = & Loss_{event} + Loss_{corefer} + Loss_{temp}\\
        & + Loss_{causal} + Loss_{subevent}
    \end{split}
\end{equation}

\subsection{Event Relation Graph Encoder}


The event relation graph encoder is designed to encode the event relation graph based on hard labels. In this process, event embeddings are aggregated with neighbor events embeddings through interconnected relations, and sentence embeddings are updated with events and event relations embeddings.

To capture sentence-level context and explicitly demonstrate the connection between each sentence and its reported events, we introduce an extra node to represent each sentence and connect it with the associated events, as depicted in Figure \ref{methodology_figure}.

The resulting event relation graph contains two types of nodes (event nodes and sentence nodes) and nine types of heterogeneous relations (\textit{coreference}, \textit{before}, \textit{after}, \textit{overlap}, \textit{causes}, \textit{caused by}, \textit{contains}, \textit{contained by} relations between events, and event-sentence relation). The event node is initialized as the event word embedding, and the sentence node is initialized as the embedding of the sentence start token <s>. The eight event-event relations are constructed based on hard labels and inherently carry semantic meaning. In order to integrate the semantic meaning of event relations into graph propagation, we introduce the relation-aware graph attention network. In terms of event-sentence relation which is a standard link without semantics, we utilize the standard graph attention network \cite{veličković2018graph}.

The relation-aware graph attention network processes event-event relations, and propagates relations semantic meaning into the connected events embeddings. The event-event relation $r_{ij}$ connecting $i$-th and $j$-th event nodes is initialized as the embedding of the corresponding relation word $r$. During the propagation at the $l$-th layer, the input of $i$-th event node is the output produced by the previous layer denoted as $e_i^{(l-1)}$, and the relation embedding $r_{ij}$ is updated as:
\begin{equation}
    r_{ij} = W^r[e_i^{(l-1)} \oplus r_{ij} \oplus e_j^{(l-1)}]
\end{equation}
where $\oplus$ represents feature concatenation and $W^r$ is a trainable matrix. Then the attention weights across neighbor event nodes are computed as:
\begin{equation}
    \alpha_{ij} = softmax_j\big((W^Qe_i^{(l-1)})(W^Kr_{ij})^T\big)
\end{equation}
where $W^Q$, $W^K$ are trainable matrices. The output feature for $i$-th event node regarding the relation type $r$ is formulated as:
\begin{equation}
    e_{i, r}^{(l)} = \sum_{j\in \mathcal{N}_{i,r}}\alpha_{ij}W^Vr_{ij}
\end{equation}
where $W^V$ is a parameter, and $\mathcal{N}_{i,r}$ denotes the neighbor event nodes connecting with $i$-th event via the relation type $r$. The same procedure to derive the $i$-th event embedding for all the relation types \small$r \in R = $ \{\textit{coreference}, \textit{before}, \textit{after}, \textit{overlap}, \textit{causes}, \textit{caused by}, \textit{contains}, \textit{contained by}\} \normalsize. The final output feature for $i$-th event node at the $l$-th layer is accumulated as:
\begin{equation}
    e_i^{(l)} = \sum_{r\in R}e_{i, r}^{(l)}/|R|
\end{equation}

The standard graph attention network handles the event-sentence relation, and updates sentence node embeddings with interconnected events and event relations embeddings. During the propagation at $l$-th layer, the input of $k$-th sentence node is the output feature produced by the previous layer denoted as $s_k^{(l-1)}$, and the attention weights across the connected events are:
\begin{equation}
    \small
    h_{kj} = LeakyReLU\big(a^T\big[Ws_k^{(l-1)}\oplus We_j^{(l-1)}\big]\big)
\end{equation}
\begin{equation}
    \small
    \alpha_{kj} = softmax_j(h_{kj}) = \frac{\exp(h_{kj})}{\sum_{j\in \mathcal{N}_{k}}\exp(h_{kj})}
\end{equation}
where $\mathcal{N}_{k}$ is the set of event nodes reported in $k$-th sentence. The final output feature for $k$-th sentence at the $l$-th layer is calculated as:
\begin{equation}
    s_k^{(l)} = \sum_{j\in \mathcal{N}_{k}}\alpha_{kj}We_j^{(l-1)}
\end{equation}
The sentence node embedding generated at the last layer captures the features of interconnected events and event relations, encompassing both the graph structure and sentence context. We further build a two-layer classification head on top to predict whether the sentence contains bias. The classical cross entropy loss is used for training.

\section{Experiments}

\subsection{Datasets}

Among the existing datasets, BASIL \cite{fan-etal-2019-plain} and BiasedSents \cite{lim-etal-2020-annotating} are the only two datasets that annotate sentence-level media bias while considering the article context. Other studies \cite{DBLP:journals/corr/abs-2105-11910, spinde2021neural} annotated bias sentences individually without taking into account the broader context. Recognizing that media bias can be very subtle and usually depends on comprehensive contextual information to be discovered \cite{fan-etal-2019-plain}, we utilize BASIL and BiasedSents datasets in the subsequent experiments. Table \ref{dataset_statistics} shows the statistics of the two datasets.

\begin{itemize}
    \item \textbf{BASIL} dataset \cite{fan-etal-2019-plain} gathered 300 articles from the period of 2010 to 2019. Both lexical bias and informational bias are annotated: lexical bias shifts public opinions through overt ideological language, while informational bias selectively includes or omits information. Because both types of bias can introduce ideological bias to the readers and sway their opinions, we consider them both in our bias sentences identification task. To be specific, we label a sentence as \textit{bias} if it carries either type of bias, or assign the \textit{non-bias} label if neither type of bias exists.
    \item \textbf{BiasedSents} \cite{lim-etal-2020-annotating} collected 46 articles from the year of 2017 to 2018. Each sentence is annotated into four scales: not biased, slightly biased, biased, and very biased. Following the previous work on binary judgments \cite{fan-etal-2019-plain, lei-etal-2022-sentence}, we also process the first two scales as \textit{non-bias} class and the latter two as \textit{bias} class. The dataset releases the annotations from five different annotators, from which we derive the majority voting label as the ground truth.
\end{itemize}

\begin{table}[t]
    \centering
    \scalebox{0.9}{\begin{tabular}{|l|c|c|c|c|}
        \hline
        Dataset & \# Article & \# Sent & \# Bias & \% Bias \\
        \hline
        BASIL & 300 & 7977 & 1623 & 20.34 \\
        \hline
        BiasedSents & 46 & 842 & 290 & 34.44 \\
        \hline
    \end{tabular}}
    \caption{Number of articles, sentences, bias sentences, and the ratio of bias sentences in the two datasets.}
    \label{dataset_statistics}
\end{table}

\subsection{Evaluation of Event Relation Graph}

The event relation graph is trained on the recent MAVEN-ERE dataset \cite{wang-etal-2022-maven} which annotates all the four event relations within a large scale of general-domain news articles. We employ the current state-of-art model framework \cite{wang-etal-2022-maven} to train different components collaboratively. Table \ref{event_identification_result} presents the performance of event identification. Table \ref{coreference_result} shows the performance of event coreference resolution. Following the previous work \cite{cai-strube-2010-evaluation}, MUC \cite{vilain-etal-1995-model}, $B^3$ \cite{10.3115/980845.980859}, $CEAF_e$ \cite{luo-2005-coreference}, and BLANC \cite{recasens_hovy_2011} are used as evaluation metrics. The performances of other components in the event relation graph, including temporal, causal, and subevent relation classification are summarized in Table \ref{temp_causal_subevent_result}. The standard macro-average precision, recall, and F1 score are reported.

\begin{table*}[ht]
    \centering
    \scalebox{0.95}{
    \begin{tabular}{|l||ccc||ccc|}
        \hline
        & \multicolumn{3}{c||}{BASIL} & \multicolumn{3}{c|}{BiasedSents} \\
        \hline
        & Precision & Recall & F1 & Precision & Recall & F1 \\
        \hline
        Baseline Model & & & & & & \\
        all-bias & 20.34 & 100.00 & 33.81 & 34.44 & 100.00 & 51.23 \\
        gpt-3.5-turbo & 32.22 & 42.39 & 36.61 & 42.22 & 30.25 & 35.25 \\
        gpt-3.5-turbo + 5-shot & 33.57 & 43.07 & 37.73 & 42.08 & 32.16 & 36.46 \\
        gpt-3.5-turbo + full article & 37.58 & 44.45 & 40.73 & 40.15 & 34.39 & 37.05 \\
        gpt-3.5-turbo + event relation graph & 32.34 & 52.86 & 40.13 & 40.93 & 48.06 & 44.21 \\
        \cite{lei-etal-2022-sentence} & 49.41 & 48.80 & 49.10 & 42.81 & 83.44 & 56.59 \\
        longformer & 46.81 & 45.65 & 46.22 & 39.78 & 79.30 & 52.98 \\
        longformer + additional features & 47.02 & 46.33 & 46.67 & 40.27 & 79.65 & 53.49 \\
        \hline
        Event Relation Graph & & & & & & \\
        + event-aware language model (soft label) & 50.97 & 48.61 & 49.76 & 46.47 & 79.62 & 58.68 \\
        + event relation graph encoder (hard label) & 47.28 & 53.11 & 50.03 & 49.72 & 84.39 & 62.57 \\
        + both (full model) & \textbf{50.06} & \textbf{54.10} & \textbf{52.00} & \textbf{54.10} & \textbf{84.08} & \textbf{65.84} \\
        \hline
    \end{tabular}}
    \caption{Performance of sentence-level media bias identification based on event relation graph. Precision, Recall, and F1 score of the \textit{bias} class are shown. The model with the best performance is \textbf{bold}.}
    \label{media_bias_result}
\end{table*}

\subsection{Experimental Settings}

In the experiments, the model takes the entire news article as input, and outputs the prediction for each sentence within the article. Follow the previous work \cite{fan-etal-2019-plain, lei-etal-2022-sentence}, we also perform ten-folder cross validation. Instead of splitting the dataset into ten folders based on individual sentences like \citet{fan-etal-2019-plain} did, we follow \citet{lei-etal-2022-sentence} to divide the ten folders based on articles. This ensures sentences from the same article do not appear in both the training and testing folds, thus preventing knowledge leakage. In each iteration, one fold is designated as the test set, eight folds are utilized as the training set, and the remaining fold is the validation set to determine the stopping point for training. The maximum training epochs for each iteration is set to 5. We utilize the AdamW optimizer \cite{loshchilov2019decoupled}, with a linear scheduler to adaptively adjust the learning rate. The weight decay is set to 1e-2. Upon collecting the prediction results for the ten testing folds, precision, recall, and F1 score of the \textit{bias} class are calculated for evaluation.

\begin{figure*}[ht]
  \centering
  \includegraphics[width = 6.3in]{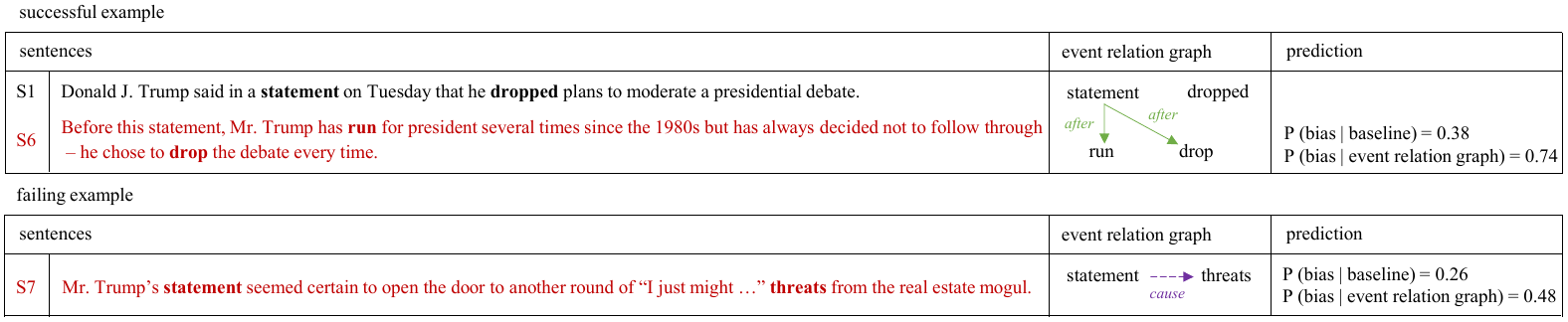}
  \caption{Example of our method succeed in solving false negative error, and a failing example. Bias sentences are highlighted in \textcolor{red}{red}. Events words are shown in \textbf{bold} text. The solid arrows in the event relation graphs represent the successfully extracted event relations, and the dashed arrow means the missing event relation.}
  \label{analysis_example}
\end{figure*}

\subsection{Baselines}

Sentence-level media bias has a relatively short research history, and there are only a few established methods available for comparison. The following systems are implemented as baselines:
\begin{itemize}
    \item \textbf{all-bias}: a naive baseline that predicts all the sentences into the \textit{bias} class
    \item \textbf{gpt-3.5-turbo}: an instruction prompt (Appendix \ref{prompt}) is crafted to enable the large language model gpt-3.5-turbo to automatically generate the predicted labels
    \item \textbf{gpt-3.5-turbo + 5-shot}: provide five \textit{bias} class examples and five \textit{non-bias} class examples as demonstrations into the gpt-3.5-turbo prompt
    \item \textbf{gpt-3.5-turbo + full article}: provide the full article context into the gpt-3.5-turbo prompt
    \item \textbf{gpt-3.5-turbo + event relation graph}: guide the large language model gpt-3.5-turbo to reason the event relation graph of the article, and then make the prediction for each sentence through a chain-of-thought process \cite{wei2023chainofthought} (Appendix \ref{prompt})
    \item \citet{lei-etal-2022-sentence}: our previous work that incorporates news discourse structures and discourse relations between sentences for bias sentence identification. For fair comparison, we update the language model used in that work into Longformer \cite{beltagy2020longformer} and include sentences that contain either lexical bias or informational bias as \textit{bias} sentences in the BASIL dataset. This ensures that both the language model and dataset processing are the same across models.
    \item \textbf{longformer}: the same language model Longformer \cite{beltagy2020longformer} is used to encode the article and an extra layer of Bi-LSTM \cite{huang2015bidirectional} is added on top. The hidden state at the sentence start token is extracted as the sentence embedding. This baseline model is equivalent to our developed model without the event relation graph.
    \item \textbf{longformer + additional features}: concatenates the probabilistic vector eq (1)-(5) as additional features with the sentence embedding.
\end{itemize}

\subsection{Experimental Results}

The experimental results of bias sentences identification based on event relation graph are presented in Table \ref{media_bias_result}. The last row is the performance of our developed model by leveraging both soft labels and hard labels within the event relation graph.

The results demonstrate that incorporating the event relation graph can significantly improve both precision and recall. Compared to the longformer baseline that lacks event relation graph, our proposed method can improve both precision and recall for the BASIL and BiasedSents datasets. This indicates that the event relation graph captures contextual knowledge and facilitates a more comprehensive understanding of the broader discourse.

Moreover, the event relation graph method performs better than the simple feature concatenation. The model that concatenates probabilistic features (longformer + additional features) exhibits only marginal performance improvements over the longformer baseline. This highlights the necessity to develop more sophisticated model to fully harness the information within the event relation graph.

Furthermore, the event relation graph that constructs event-level content structure outperforms the method based on sentence-level relations. Compared to the method incorporating sentence relations \cite{lei-etal-2022-sentence}, leveraging the event relation graph leads to superior performance. One possible explanation is that our event relation graph interprets news narratives at a more fine-grained event level, and captures event-level discourse relations across the entire article. This capability enables a more detailed comprehension of each sentence and a deeper understanding of the overall context.

In addition, our method based on event relation graph outperforms the gpt-3.5-turbo baselines. We observe that gpt-3.5-turbo baselines always choose \textit{bias} label when the text is explicitly sentimental. However, the sentence-level ideological bias can be very subtle and usually expressed in a neutral tone \cite{van-den-berg-markert-2020-context}. This demonstrates that encoding broader context is necessary to discover such implicit ideological bias. Besides, we observe that guiding the large language model to reason the event relation graph (gpt-3.5-turbo + event relation graph) significantly increases the Recall of bias sentences. By leveraging the knowledge from the event relation graph, the large language model gains better understanding of the article context, thereby uncovering more implicit bias cases.

\begin{table}[t]
    \centering
    \scalebox{0.9}{
    \begin{tabular}{|l|ccc|}
        \hline
        & Precision & Recall & F1 \\
        \hline
        baseline & 46.81 & 45.65 & 46.22 \\
        full model & \textbf{50.06} & \textbf{54.10} & \textbf{52.00} \\
        - coreference & 48.36 & 53.48 & 50.79 \\
        - temporal & 48.56 & 52.98 & 50.68 \\
        - causal & 49.91 & 50.71 & 50.30 \\
        - subevent & 48.65 & 52.37 & 50.45 \\
        \hline
    \end{tabular}}
    \caption{Effect of removing each of the four event relations. Take BASIL dataset as an example.}
    \label{effect_of_four_relations}
\end{table}

\subsection{Ablation Study}

The ablation study of leveraging either soft labels or hard labels are reported in Table \ref{media_bias_result}. The utilization of soft labels to train an event-aware language model enhances performances on both datasets, especially precision. This indicates that soft labels guide the model to learn nuanced probabilistic information from the event relation graph, enabling it to make more precise predictions. The employment of hard labels to build a heterogeneous graph attention network also improves metrics on both datasets, especially recall. This implies that hard labels facilitate propagating sentence embeddings with interconnected events embeddings, thereby encourage the encoding of a broader contextual knowledge and enable recalling more implicit cases. Incorporating them both exhibit the best performance, demonstrating that soft and hard labels are complementary with each other.

\subsection{Effect of Four Event Relations}

We further study the effect of four event relations. The experimental results of removing each of the four event relations from the full model are shown in Table \ref{effect_of_four_relations}. The results indicate that removing any one of the four relations leads to a performance decrease in both precision and recall. The four commonly-seen event relations are essential in constructing a unified and cohesive content structure. Incorporating all the four relations in the event relation graph yields the best performance.

\subsection{Analysis and Discussion}

Figure \ref{analysis_example} shows a successful example in which the event relation graph effectively recognizes the temporal relation. As a result, the model is aware that the sentence (S6) intends to disclose the past habitual behaviors of Trump, and successfully recalls it as \textit{bias} class. The current trained event relation graph has the ability to extract most easy cases through explicit discourse connectives such as \textit{before} or \textit{since} in this example.

Figure \ref{analysis_example} also shows a failing example, where the event relation graph fails to recognize the causal relation and leads to a false negative error. It is because the text expresses the causal relation in an implicit way by using the phrase \textit{open the door to}. Such implicit cases that lack explicit cues pose challenges for the current event relation graph. To further improve the performance of bias sentences identification, it is necessary to enhance the extraction of implicit event relations.

\section{Conclusion}

This paper identifies media bias at the sentence level, and demonstrates the pivotal role of events and event-event relations in identifying bias sentences. We observe that interpreting events in association with other events in a document is critical in identifying bias sentences. Inspired by this observation, this paper proposes to construct an event relation graph as the extra guidance for bias sentence detection. To integrate the event relation graph, we design a novel framework that consists of two consecutive steps to first update the language model and then further update sentence embeddings, by leveraging the soft and hard event relation labels respectively from the event relation graph. Experimental results demonstrate the effectiveness of our approach. For future work, we will further improve the extraction of implicit event relations as well as develop new interpretable methods that leverage event relation graph for media bias analysis.


\section*{Limitations}

Our paper propose to construct an event relation graph for each article to facilitate ideological bias identification. The current state-of-art model framework is employed for training the event relation graph. While the current event relation graph successfully extracts event relations for most easy cases with explicit discourse connectives or language cues, it may encounter challenges in recognizing implicitly stated event relations. Thus, improving event relation graph construction becomes necessary and serves as the future work.

\section*{Ethical Considerations}

This paper is a research paper on identifying media bias. The goal of this research is to understand, detect, and mitigate the political bias. The examples in this paper are only used for research purpose, and do not represent any political leaning of the authors. The release of the datasets and code should be used for mitigating the political bias, instead of expanding or disseminating the political bias.

\section*{Acknowledgements}

We would like to thank the anonymous reviewers for their valuable feedback and input. We gratefully acknowledge support from National Science Foundation via the awards IIS-1942918 and IIS-2127746. Portions of this research were conducted with the advanced computing resources provided by Texas A\&M High-Performance Research Computing.

\bibliography{anthology,custom}

\appendix

\section{Prompt for gpt-3.5-turbo baselines}

\label{prompt}

The instruction prompt for gpt-3.5-turbo baseline that takes individual sentences as input is: "Biased sentence with political bias refers to the supportive or background content to sway readers opinions in an ideological direction, either through information selective inclusion or overt ideological language. Please reply "Yes" if the following sentence is a biased sentence with political bias, else reply "No". Sentence: "<sentence>" Answer:"

The instruction prompt for gpt-3.5-turbo + full article baseline that takes the full article context as input is: "Biased sentence with political bias refers to the supportive or background content to sway readers opinions in an ideological direction, either through information selective inclusion or overt ideological language. Given the article: "<article>". Please reply "Yes" if the following sentence is a biased sentence with political bias, else reply "No". Sentence: "<sentence>" Answer:"

The instruction prompt for gpt-3.5-turbo + event relation graph baseline that guides the model to reason the event relation graph of the article and then make the prediction for each sentence through a chain-of-through process is: "Biased sentence with political bias refers to the supportive or background content to sway readers opinions in an ideological direction, either through information selective inclusion or overt ideological language. The task is identifying whether a sentence is biased or not. Let’s think step by step. Firstly, explain the events reported in the sentence and the event relations in the article context. Events refer to an occurrence or action reported in the sentence. There are four types of relations between events: coreference, temporal, causal, and subevent relations. Coreference relation represents whether two event mentions designate the same occurrence. Temporal relation represents the chronological order between events, such as before, after, and overlap. Causal relation represents the causality between events. Subevent relation represents the containment relation from a parent event to a child event. Secondly, answer "Yes" if the following sentence is a biased sentence with political bias, else reply "No". Please mimic the output style in the following example.\\ Article: "Donald J. Trump said in a statement on Tuesday that he dropped plans to moderate a presidential debate. Trump said that he dropped because the Republican Party asked him to give up the right to run as an independent candidate if he moderates the debate. "It is very important to me that the right Republican candidate be chosen to defeat the failed Obama administration," Mr. Trump said in his statement. "But if that Republican, in my opinion, is not the right candidate, I am not willing to give up my right to run as an independent candidate." Mr. Trump drew considerable criticism from the Republican establishment and attracted the ridicule from the public for his decision to drop and his willingness to still run if he was displeased with the candidates. Before this statement, Mr. Trump has run for president several times since the 1980s but has always decided not to follow through – he chose to drop the debate every time."\\ Sentence: "Before this statement, Mr. Trump has run for president several times since the 1980s but has always decided not to follow through – he chose to drop the debate every time."\\ Output: Firstly, explain the events reported in the sentence and the event relations in the article context. This sentence reports the events: run for president, drop the debate. The event relations in the article context: the event "run for president" is a historical event that happened before the event "statement". Trump claimed a causal relation between the event "drop the debate" and the event "give up right as an independent candidate". This sentence reports a historical event to reveal Trump’s consistent past behavior of dropping presidential election several times since 1980s, and hints that the claimed cause by Trump is not his real reason to drop the debate. Therefore, this sentence contains political bias to sway readers’ opinion towards Trump to a negative direction. Secondly, answer "Yes" if the following sentence is a biased sentence with political bias, else reply "No". Answer: Yes.\\ Article: "<article>"\\ Sentence: "<sentence>"\\ Output:"

\section{Evaluation of Event Relation Graph}

This section reports the performance of the event relation graph. The event relation graph is trained on the recent MAVEN-ERE dataset \cite{wang-etal-2022-maven} which annotates all the four event relations within a large scale of general-domain news articles. Since events and four event relations interact with each other to form a cohesive narrative framework, we adopt joint learning framework \cite{wang-etal-2022-maven} to train different components collaboratively. The AdamW \cite{loshchilov2019decoupled} is used as the optimizer. The training epochs is 10. The learning rate is initialized as 1e-5 and adjusted by a linear scheduler. The weight decay is set to 1e-2.

Table \ref{event_identification_result} presents the performance of event identification. Table \ref{coreference_result} shows the performance of event coreference resolution. Following the previous work \cite{cai-strube-2010-evaluation}, MUC \cite{vilain-etal-1995-model}, $B^3$ \cite{10.3115/980845.980859}, $CEAF_e$ \cite{luo-2005-coreference}, and BLANC \cite{recasens_hovy_2011} are used as evaluation metrics. The performances of other components in the event relation graph, including temporal, causal, and subevent relation classification are summarized in Table \ref{temp_causal_subevent_result}. The standard macro-average precision, recall, and F1 score are reported.

\begin{table*}[ht]
    \centering
    \scalebox{1.0}{
    \begin{tabular}{|c|ccc|}
        \hline
        & Precision & Recall & F1 \\
        \hline
        Event Identifier & 87.31 & 91.81 & 89.40 \\
        \hline
    \end{tabular}}
    \caption{Performance of event identification. Macro precision, recall, and F1 are reported.}
    \label{event_identification_result}
\end{table*}

\begin{table*}[ht]
    \centering
    \scalebox{0.87}{
    \begin{tabular}{|ccc|ccc|ccc|ccc|}
        \hline
        \multicolumn{3}{|c|}{MUC} & \multicolumn{3}{|c|}{$B^3$} & \multicolumn{3}{|c|}{$CEAF_e$} & \multicolumn{3}{|c|}{BLANC} \\
        \hline
        Precision & Recall & F1 & Precision & Recall & F1 & Precision & Recall & F1 & Precision & Recall & F1 \\
        \hline
        76.34 & 83.10 & 79.57 & 97.07 & 98.32 & 97.69 & 97.79 & 97.00 & 97.39 & 83.69 & 92.43 & 87.54 \\
        \hline
    \end{tabular}}
    \caption{Performance of event coreference resolution in the event relation graph}
    \label{coreference_result}
\end{table*}

\begin{table*}[ht]
    \centering
    \scalebox{1.0}{
    \begin{tabular}{|c|ccc|}
        \hline
        & Precision & Recall & F1 \\
        \hline
        Temporal & 48.45 & 46.43 & 47.04 \\
        Causal & 58.48 & 54.02 & 56.01 \\
        Subevent & 53.37 & 42.90 & 46.21 \\
        \hline
    \end{tabular}}
    \caption{Performance of temporal, causal, and subevent relation tasks in the event relation graph. Macro precision, recall, and F1 are reported.}
    \label{temp_causal_subevent_result}
\end{table*}

\end{document}